\documentclass[runningheads]{llncs}

\usepackage{eccv}

\usepackage{eccvabbrv}

\usepackage{graphicx}
\usepackage{booktabs}

\usepackage{amsmath}
\usepackage{multirow,multicol}
\usepackage{color}
\usepackage[accsupp]{axessibility}  

\usepackage[pagebackref,breaklinks,colorlinks]{hyperref}

\usepackage{orcidlink}

\begin{document}

\title{CSDNet: Detect Salient Object in Depth-Thermal via A Lightweight Cross Shallow and Deep Perception Network} 

\titlerunning{Abbreviated paper title}

\author{Xiaotong Yu\inst{1}\orcidlink{0000-1111-2222-3333} \and
Ruihan Xie\inst{2}\orcidlink{1111-2222-3333-4444} \and
Zhihe Zhao\inst{2}\orcidlink{2222-3333-4444-5555} \and
Chang-Wen Chen\inst{1}\orcidlink{3333-4444-5555-6666}} 

\authorrunning{F.~Author \etal}

\institute{The Hong Kong Polytechnic University, HKSAR \and
The Chinese University of Hong Kong, HKSAR}

\maketitle

\begin{abstract}
While we enjoy the richness and informativeness of multimodal data, it also introduces interference and redundancy of information. To achieve optimal domain interpretation with limited resources, we propose CSDNet, a lightweight \textbf{C}ross \textbf{S}hallow and \textbf{D}eep Perception \textbf{Net}work designed to integrate two modalities with less coherence, thereby discarding redundant information or even modality. We implement our CSDNet for Salient Object Detection (SOD) task in robotic perception. The proposed method capitalises on spatial information prescreening and implicit coherence navigation across shallow and deep layers of the depth-thermal (D-T) modality, prioritising integration over fusion to maximise the scene interpretation. To further refine the descriptive capabilities of the encoder for the less-known D-T modalities, we also propose SAMAEP to guide an effective feature mapping to the generalised feature space. Our approach is tested on the VDT-2048 dataset, leveraging the D-T modality outperforms those of SOTA methods using RGB-T or RGB-D modalities for the first time, achieves comparable performance with the RGB-D-T triple-modality benchmark method with 5.97 times faster at runtime and demanding 0.0036 times fewer FLOPs. Demonstrates the proposed CSDNet effectively integrates the information from the D-T modality. The code will be released upon acceptance. 
  
\end{abstract}

\section{Introduction}
\label{sec:intro}

In recent decades, various multimodality techniques have shown significant advancements in this learning era, especially in the field of robotic perception. Within the realm of multimodality techniques, it is commonly observed that the performance of models tends to improve with an increasing number of modalities \cite{huang2021makes}. However, this inevitably leads to higher costs, higher computational demands, and unpredictable noise. In contrast, crossing off a modality without proper rationale may also cause the loss of crucial scene interpretation \cite{wan2023mffnet}, presenting a trade-off dilemma. Nevertheless, we identify an entirely new pivot beyond this lever: the integration of modalities with low coherence to achieve broader domain coverage, exploring the possibility of breaking the constraint of 'adding modalities for better interpretation'. 
\begin{figure}[!ht]
    \centering
    \includegraphics[width=12cm]{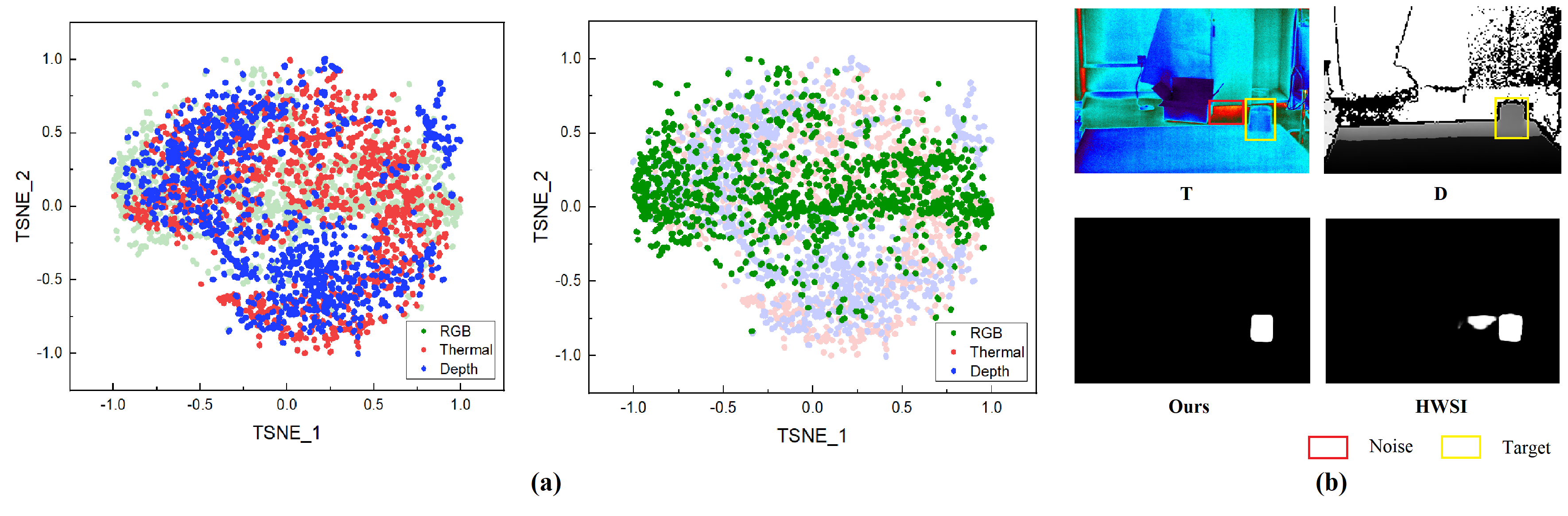}
    \caption{(a) The TSNE representations of different modalities; (left) Depth and thermal are highlighted; (right) RGB modality is highlighted (b) The visualised results of existing methods on D-T modality, the RGB-dominated models show less capability in interpreting D-T data.}
    \label{fig:cover}
\end{figure}

In the existing multimodal studies, RGB typically serves as the primary modality with high priority since it is widely recognised as a senior modality due to its rich texture information, colour and high-resolution spatial details. Depth and thermal modalities are considered as subsidiary modalities since they provide relatively limited but more specialised information with particular significance. Conventional multimodal approaches on visual perception tasks by RGB-D \cite{zhang2021rgb, zhou2021specificity, zhang2023feature, zhao2022self}, RGB-T \cite{zhou2023position, liu2023scribble, xie2023cross, zhou2023lsnet} and RGB-D-T \cite{song2022novel, wan2023mffnet} have already achieved commendable results. With the addition of depth information (\ie, RGB-D) the dual modality model facilitated a more comprehensive understanding of 3D geometry, proving inherently superior to monochrome RGB, particularly in challenging environments such as low light \cite{feng2019attentive}. Additionally, the inclusion of thermal images accentuates specific temperature variances within the sensing domain, reducing distractions and enhancing domain awareness under clustered environments or challenging lighting conditions. In this case, triple-modality data (RGB-depth-thermal) could potentially offer the most comprehensive information. Therefore, the triple-modality benchmark method \cite{song2022novel} yields superior results compared to the other two multimodal combinations (RGB-D and RGB-T) \cite{song2022novel, wan2023mffnet} at the cost of a substantial model size of 403.4 MB and a high computational complexity of 357.69G FLOPs. Indeed, in contrast to RGB-D-T methods, neither RGB-D nor RGB-T approaches can offer a complete representation. However, indiscriminately incorporating additional modalities in multimodal tasks can lead to data redundancy, increased computational costs, and potential privacy concerns. In order to identify the redundant elements across these three modalities, we analyse the intrinsic characteristics of different modalities, revealing significant overlap between RGB and depth or thermal, while depth and thermal exhibit less overlap but offer broad domain coverage, as shown in Figure \ref{fig:cover}(a) indicating that D-T introduces less redundancy. Nevertheless, unlike the texture coherence of RGB-T and the spatial coherence of RGB-D, the lack of coherence between depth and thermal modalities poses a challenge for existing multimodal methods when dealing with depth-thermal data. Some visualisation results, as shown in Figure \ref{fig:cover}(b), indicate significant discrepancies in the description of the same region in D-T. Combining these two modalities with low coherence can confuse the model in making incorrect interpretations. To address the challenge of the information integration of the low coherence modalities, we present the Cross Shallow and Deep Perception Network, CSDNet. The cross shallow and deep scheme involves two stages, we explore the capacity of synergies between the two modalities by crossing the saliency-aware prescreening mask at the shallow layer and, meanwhile, introducing an implicit coherence activation operation at the deep layer to reasonably select the similarities and distinctions between high-level features derived from these disparate inputs. Furthermore, considering that the selected backbone of our encoder, MobileNet-V2, was pretrained on the expansive RGB dataset ImageNet, which shows less interpretation of depth and thermal, we utilise the powerful and robust Segment Anything Model (SAM) to guide the encoder in mapping the D-T into a generalised feature space. To the best of our knowledge, this is not only the first study to investigate the low coherence modality synergies using depth and thermal data, but also the first study of applying D-T modalities to the task of salient object detection. 

The main contribution of this work can be summarised as follows:
\begin{itemize}
    \item We propose a novel cross shallow and deep perception scheme to maximise the scene interpretation by leveraging low-coherence modalities. 
    \item We introduce an innovative SAM-assist encoder pre-training framework to guide the encoder to extract more generalised features. 
    \item We implement the proposed method to the salient object detection network using only depth and thermal images. Comprehensive experiments are carried out to demonstrate the effective integration of depth and thermal modalities. 
\end{itemize}




\section{Related Works}

\subsection{Dual-modal and Triple-modal Salient Object Detection}

The progression of image capture technologies in recent years has catalysed the integration of depth and thermal imaging within Salient Object Detection (SOD) tasks. In the realm of depth information, Huang \etal \cite{huang2022middle} introduced an RGB-D saliency detection model that utilises dual shallow subnetworks for the extraction of unimodal RGB and Depth features at varying tiers. Concurrently, In \cite{song2022improving}, Song \etal presented a modality-aware decoder, which encompasses feature embedding, modality reasoning, and strategies for feature back-projection and collection. Furthermore, Bi \etal \cite{bi2023cross} advanced a cross-modal hierarchical interaction network, excavating and progressively fusing multi-level features. 
The positional dependence of depth information often confounds object identification, particularly for targets adjacent to their background. This limitation inherent to RGB-D-based salient object detection methodologies has prompted the incorporation of thermal imaging as an alternative input data to augment the saliency detection capabilities. Chen \etal \cite{chen2022cgmdrnet} suggested a network guided by the reduction of modality discrepancies, integrating modules that minimise the bimodal disparities, including a modality difference reduction module, cross-attention fusion module, and a transformer-based feature enhancement module.
In the existing dual-modality approaches, RGB is positioned as the primary modality, with a subsidiary modality added to augment scene interpretation. However, neither RGB-D nor RGB-T can offer a comprehensive representation compared to the triple-modality RGB-D-T. 

There are limited studies regarding the RGB-D-T salient object detection, owing to the high cost associated with the collection and alignment of triple-modality data. In \cite{song2022novel}, Song \etal introduced a pioneering visible-depth-thermal (VDT) image dataset VDT-2048 tailored for salient object detection. This dataset comprises 2048 image groups captured in 14 challenging scenes, providing a diverse range of scenarios to evaluate the generalisation performance of multimodal SOD methods. 
This study also presented a benchmark of the VDT triple-modality SOD method, which leverages a hierarchical weighted suppress interference (HWSI) architecture to fuse features from different modalities effectively. 
Subsequently, Wan \etal \cite{wan2023mffnet} addressed the limitations of single-modal and dual-modal salient object detection methods by introducing a novel approach that integrates RGB, depth, and thermal images. The proposed MFFNet consists of a triple-modality deep fusion encoder and a progressive feature enhancement decoder to enhance the complementarity between different modalities during encoding and achieve accurate saliency information recovery during decoding. The RGB-D-T methods achieve better performance than the dual-modality methods. However, they inevitably lead to larger model size and increasing computational burden. In this study, we present the low-coherence D-T modality integration method, aiming to achieve scene interpretation comparable to the triple-modality approach. 

\subsection{Segment Anything Model and Derived Works} 

The release of the Segment Anything Model (SAM) \cite{kirillov2023segment} by Meta has gained widespread attention. SAM is built around a powerful and robust image encoder, leveraging the strengths of the Vision Transformer (ViT) architecture, coupled with a lightweight decoder that generates prompt-guided masks which work in sequence. 
SAM was trained on the SA-1B dataset, which comprises an impressive collection of over 1 billion masks on 11 million images.
The extensive training offered the model with strong generalisation ability and can be easily adapted to a range of downstream vision tasks, positioning it as a pivotal foundation model in computer vision and it is believed to be a 'GPT moment' for vision. Following the release of SAM, numerous studies have been conducted to make it more mobile-friendly \cite{zhao2023fast,zhang2023faster, zhang2023mobilesamv2}, as well as to tailor it for specific data types, such as medical imagery \cite{wu2023medical,chen2023ma,gong20233dsam}. In this work, we propose the SAMAEP, which employs the SAM to guide the encoder in effectively interpreting depth and thermal data, thereby improving model performance.

\section{Proposed Method}

\begin{figure}[!htb]
    \centering
    \includegraphics[width=10cm]{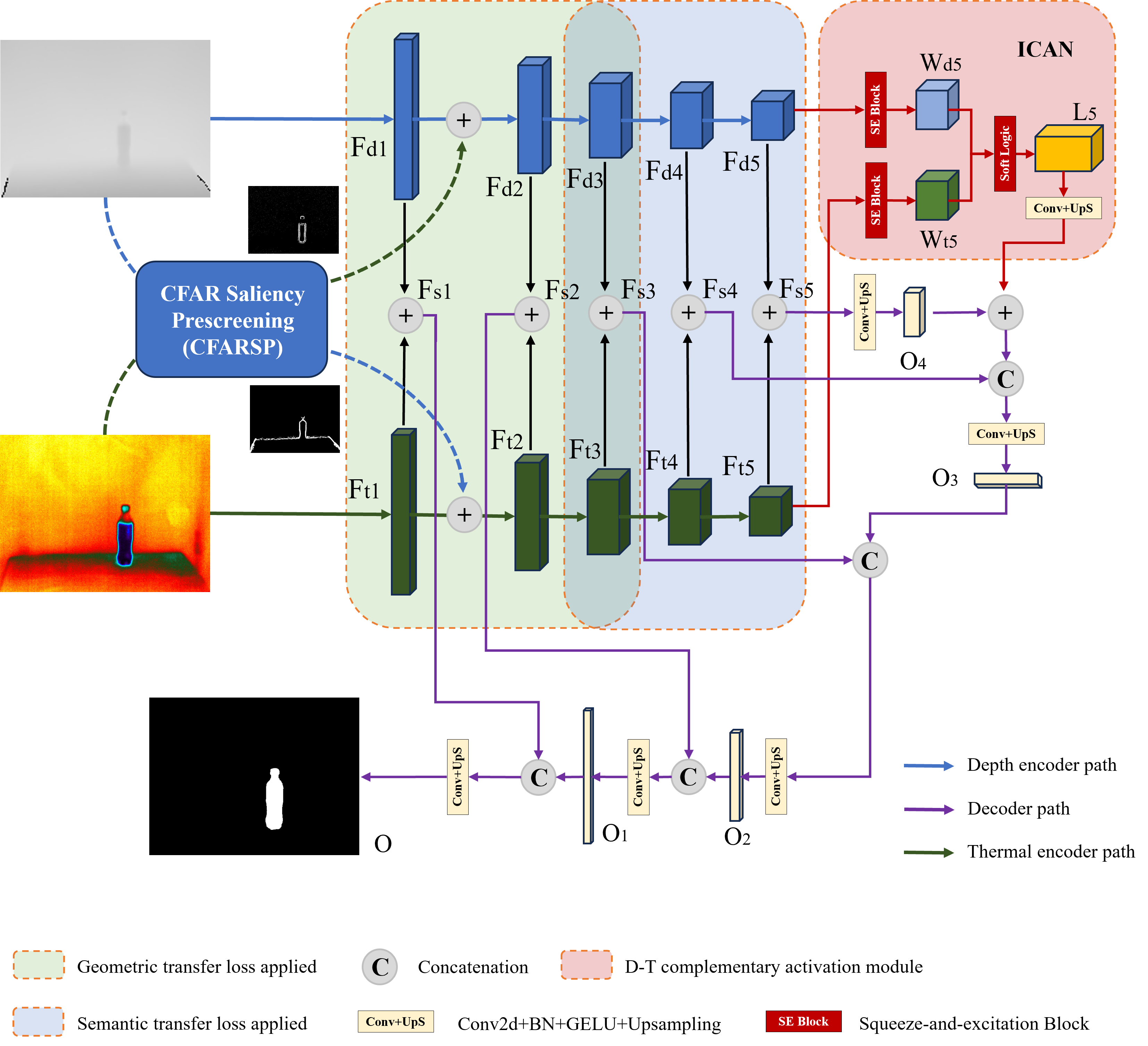}
    \caption{The overview of the proposed network CSDNet}
    \label{fig:overview}
\end{figure}

This section presents the overview of our proposed CSDNet for salient object detection relying only on thermal and depth images. For the integration of the modalities with low coherence, the prescreened spatial information is exchanged between the modalities at the shallow layer. Meanwhile, the middle layers select the relevant representations from the superposition features. And finally, the deep layer facilitates the implicit coherence between two modalities utilizing high-level features. Following this pipeline, the cross shallow and deep integration scheme consists of two modules. For the shallow layer spatial synergies, we introduce the CFAR detector-based saliency prescreening module (CFARSP), while the implicit coherence activation navigator module (ICAN) is designed for the deep layer semantic synergies. The motivation and implementation of CFARSP, ICAN and SAMAEP will be presented in \ref{CFARSP}, \ref{ICAN} and \ref{SAMAEP} in details, respectively. Finally, the loss used in the proposed method is formulated. Figure \ref{fig:overview} shows the overview of our proposed network. The overall network structure follows the conventional encoder-decoder framework. The adapted MobileNet-V2 is incorporated as the encoder backbone for both depth and thermal modalities. Representative features extracted by the encoder at five distinct scales are denoted as $F_{di}$ and $F_{ti}$, where $i=1, 2, 3, 4, 5$. The CFARSP module accepts the original depth and thermal images as inputs, yielding a saliency-aware prescreening mask. The ICAN module utilises $F_{d5}$ and $F_{t5}$ as input features and subsequently integrates the supplementary features at $O_4$. The ultimate output of the decoder is a refined saliency map, symbolised as $O$. 

\subsection{CFAR Saliency Prescreening Module}
\label{CFARSP}
\begin{figure}[!htb]
    \centering
    \includegraphics[width=8cm]{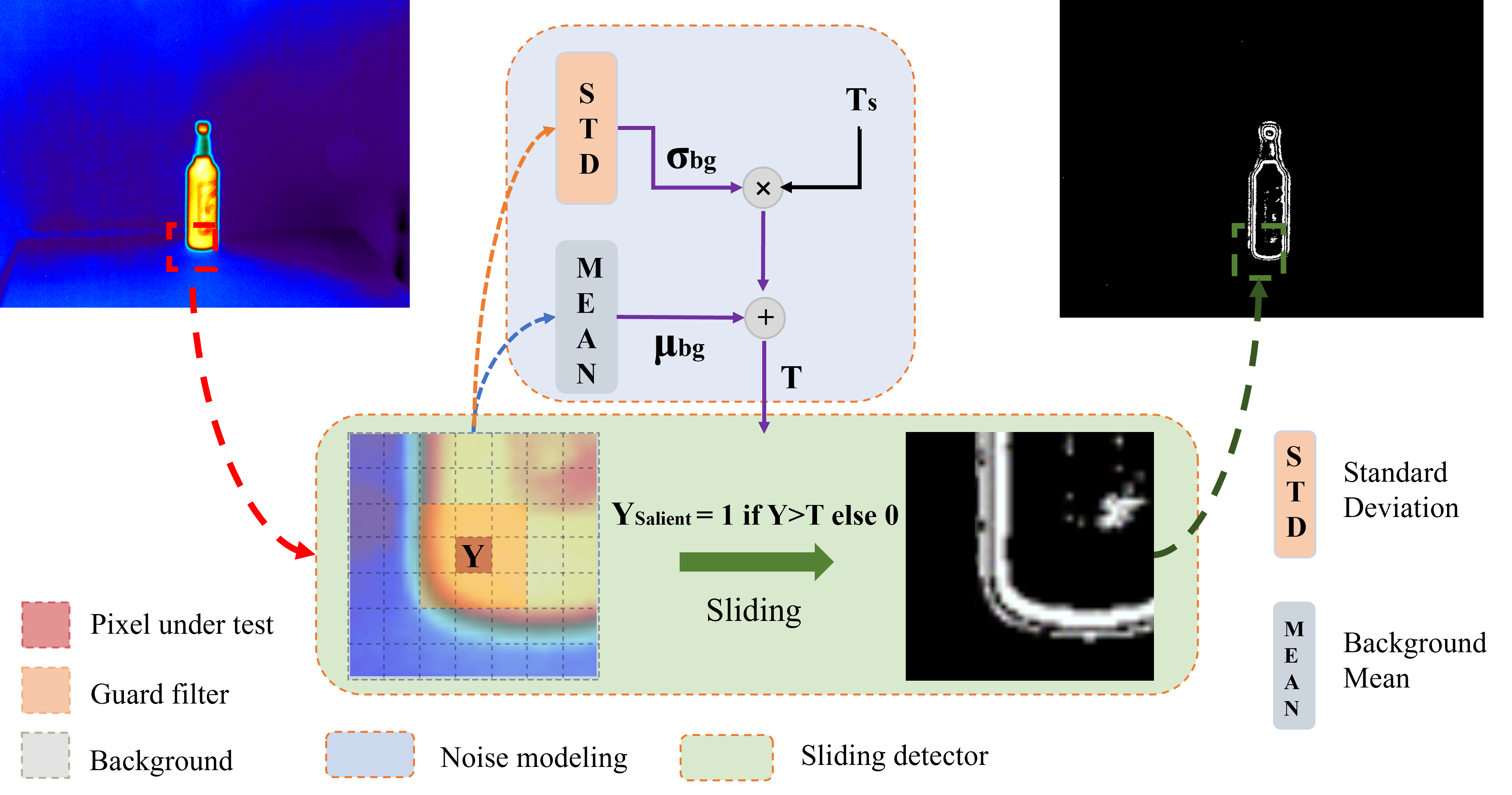}
    \caption{The schematic of CFAR Saliency Prescreening Module}
    \label{fig:cfar}
\end{figure}
As mentioned in the introduction, hastily fusing two low-coherence modalities in the shallow layers of the network can confuse the model on how to interpret them. To ensure the distinctiveness of each modality, and to exchange information between two modalities in shallow layers, we propose the CFARSP module. The overall architecture of the CFARSP module is shown in Figure \ref{fig:cfar}. The constant false alarm rate (CFAR) detection is a technique widely used in radar systems to identify and eliminate false alarm signals caused by noise or other interference. Its binary nature implies that it imposes a lower computational burden and can serve as a screening mask to highlight significant features of the modality. We follow the modelling approach of CFAR using probability density functions to describe background clutter. In this work, the 2D probability of false alarm (PFA) for the threshold $T$ can be represented as: $ PFA=1-\int_{-\infty}^T f(x)dx=\int_T^\infty f(x)dx$.

In instances of medium and lower resolution imagery, such clutter frequently adheres to a Gaussian distribution when examined within the intensity domain, or alternatively, conforms to a Rayleigh distribution within the amplitude domain, in accordance with the principles of the central limit theorem \cite{bouhlel2018maximum}. In this case, we employ the commonly used Gaussian distribution model and use a sliding window to make saliency judgments on candidate points in the background. Thus the detector can be described as: $Y>\mu_{bg}+\sigma_{bg}T_s \Leftrightarrow target$.
where $Y$ denotes the pixel under testing, $\mu_{bg}$ and $\sigma_{bg}$ represent the mean value and standard deviation of the background. And $T_s$ is the design parameter threshold scale which controls the sensitivity of CFAR detector. 

\subsection{Implicit Coherence Activation Navigation Module}
\label{ICAN}
Our analysis indicates that deep networks can have different semantic descriptions of the same scene in two modalities with lower coherence. In this context, the model can facilitate a more comprehensive scene interpretation by linking the semantic information from depth and thermal modalities wisely. Motivated by this potential for enhanced scene understanding, we introduce a new implicit coherence activation navigator aimed at activating hidden coherence relationships by emphasizing the consistency and difference between two semantics. The implementation logic is demonstrated in the upper-right corner in Figure \ref{fig:overview}. The highest level features $F_{d5}$ and $F_{t5}$ are weighted by the squeeze-and-excitation block \cite{hu2018squeeze}, denoted as $W_{d5}$ and $W_{t5}$. The soft logic operations can be represented as $AND_{w5}=min(W_{d5}, W_{t5})$, $OR_{w5}=max(W_{d5}, W_{t5})$, $XOR_{w5}=abs(W_{d5}-W_{t5})$, and then the results are concatenated along the channel axis $L_5=concat(AND_{w5}, OR_{w5}, XOR_{w5})$. The concatenated logic result is added to $O_4$ with a higher resolution feature after sequential operations of convolution, batch normalisation, Gaussian Error Linear Unit (GELU) activation and a bilinear interpolation upsampling. 


\subsection{SAM-Assist Encoder Pre-training Framework}
\label{SAMAEP}
\begin{figure}[!h]
    \centering
    \includegraphics[width=8cm]{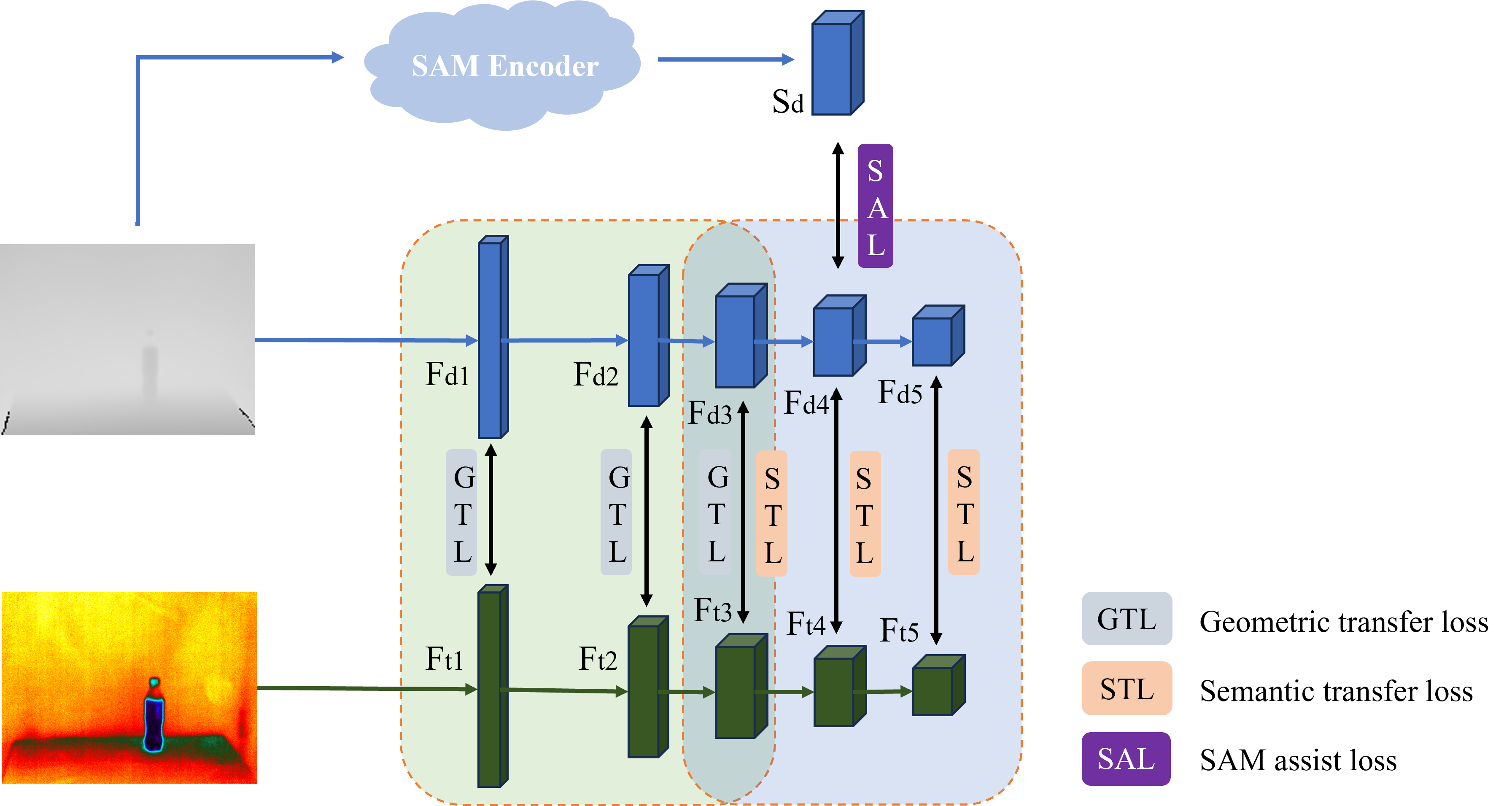}
    \caption{The schematic of SAM-assist depth encoder pre-training framework}
    \label{fig:samaep}
\end{figure}

It is noticed that the pre-trained MobileNet-V2 on the ImageNet dataset has limited capability in extracting spatial information from depth images. Informed by the recent progress in vision foundation models, notably the Segment Anything Model (SAM) \cite{kirillov2023segment}, which demonstrates exceptional aptitude in interpreting various types of images, we have incorporated a SAM-assisted depth encoder pre-training stage. This framework aims to augment the feature extraction capabilities specific to the depth modality, thereby enhancing the overall performance of the CSDNet. The schematic of the proposed SAMAEP framework is depicted in Figure \ref{fig:samaep}. The robust and powerful vision transformer (ViT)-based encoder of SAM yields an image embedding $S_d$ dimensioned at [256, 64, 64] for a single input instance. To capitalise on the potential of these embeddings, they are strategically deployed to guide the depth encoder at the fourth phase feature output using SAM-assist loss (SAL), simultaneously ensuring the full depth encoder is weakly aligned with the thermal encoder using geometric transfer loss (GTL) and semantic transfer loss (STL) proposed in \cite{zhou2023lsnet}. 

\subsection{Loss Formulation}

In the SAM-assist depth encoder pre-training stage, SAL integrate the Mean Square Error (MSE) loss with the STL. The feature from the depth encoder is denoted as $F_{di}$, and the image embedding from SAM is denoted as $S_d$. For instance, considering the STL loss transfers the feature from the $F_{di}, i=4$ to $S_d$, the channel attention for attentive transfer involves a sequence of global average pooling ($AP$), two convolutions ($Conv$) and a sigmoid activation ($Sig$), followed by a global normalisation ($GNorm$) along the channel dimension after AP. The normalized feature $w_{Fdi}$ from $F_{di}$ can be represented as: 
\begin{equation}
    w_{Fdi}=GNorm(Sig(Conv(ReLU(Conv(AP(F_{di}^{detach}))))))
\end{equation}
The L2 norm is applied on the plane dimension to preserve the representations across different channels.
\begin{equation}
    S_{d\_norm}=L2norm(S_d),\ F_{di\_norm}^{detach}=L2norm(F_{di}^{detach})
\end{equation}
Hence, the STL from the depth encoder feature to SAM embedding is represented as: 
\begin{equation}
    STL_{Fdi\rightarrow Sd}=w_{Fdi}\times MSE(F_{di\_norm}^{detach}, S_{d\_norm})
\end{equation}
The SAL is formulated as a weighted sum of MSE loss and STL, where $w_1$ and $w_2$ signify the respective weights:
\begin{equation}
    SAL=w_1\cdot MSE(F_{di}, S_d)+w_2\cdot STL_{Fdi\rightarrow Sd}
\end{equation}
Meanwhile, the GTL and STL are employed to anchor the depth encoder using the ImageNet pre-trained thermal encoder. Following the approach in \cite{zhou2023lsnet}: 
\begin{align}
    STL_{Fdi\rightarrow Fti}=\sum_{i=3}^6w_{Fdi}\times MSE(F_{di\_norm}^{detach}, F_{ti\_norm}) \\
    STL_{Fti\rightarrow Fdi}=\sum_{i=3}^6w_{Fti}\times MSE(F_{ti\_norm}^{detach}, F_{di\_norm})
\end{align}

Unlike STL, GTL calculates spatial attention rather than channel attention to obtain the global geometric weight and distinguish the significant features on the plane. The $L2norm$ in GTL is computed along the channel axis to preserve geometric information of the spatial structure. 
\begin{align}
    GTL_{Fdi\rightarrow Fti}=\sum_{i=1}^3w_{Fdi}\times MSE(F_{di\_norm}^{detach}, F_{ti\_norm}) \\
    GTL_{Fti\rightarrow Fdi}=\sum_{i=1}^3w_{Fti}\times MSE(F_{ti\_norm}^{detach}, F_{di\_norm})
\end{align}
The overall loss for the pre-training stage can be represented as follows, where $w_3$ and $w_4$ denote different weights.
\begin{equation}
\begin{split}
    L_{SAMAEP}=SAL+w_3(GTL_{Fdi\rightarrow Fti}+STL_{Fdi\rightarrow Fti}) \\
    +w_4(GTL_{Fti\rightarrow Fdi}+STL_{Fti\rightarrow Fdi})
\end{split}
\end{equation}
After the depth encoder pre-training stage, the SOD loss functions are employed for joint encoder-decoder training. The predicted saliency region should not only be accurately aligned with the ground truth but also demonstrate a good fit on the boundary of the saliency region. Therefore, the saliency region boundaries are extracted from both $GT$ and $O_i$ from the decoder. The intersection-over-union (IOU) and binary cross-entropy (BCE) are utilised to measure the accuracy and precision: 
\begin{equation}
    L_{SOD}^{O_i}=L_{ioubce}^{reg_{Oi}}+L_{ioubce}^{bou_{Oi}}
\end{equation}
Where $L_{ioubce}=L_{iou}+L_{bce}$. To achieve a better performance, the $L_{SOD}$ is calculated for the last three stages in the decoder, \ie for $O$, $O_1$ and $O_2$, with the final loss being the summation of the three terms: 
\begin{equation}
    L_{SOD} =  L_{SOD}^{O}+ L_{SOD}^{O_1}+ L_{SOD}^{O_2}
\end{equation}

\section{Experiments}

\subsection{Dataset and Evaluation Metrics}

We validate the proposed model CSDNet on the VDT-2048 dataset as reported by Song \etal \cite{song2022novel}, which contains 1048 images for training and 1000 images for testing. The performance of our model is assessed by five standard metrics for SOD. In the context of SOD evaluation, we incorporate the following five benchmark metrics: mean absolute error (MAE) \cite{perazzi2012saliency}, F-measure ($F_m$) \cite{achanta2009frequency}, weighted F-measure ($W_F$) \cite{margolin2014evaluate}, structure measure ($S_m$) \cite{fan2017structure} and E-measure ($E_m$) \cite{fan2018enhanced}: 

\textbf{(1) Mean absolute error (MAE)} quantifies the difference between the predicted saliency map $O$ and ground truth $GT$ on a pixel-by-pixel basis: $MAE=\sum_{i=1}^w\sum_{j=1}^h |O(i,j)-GT(i,j)|/(w\times h)$. 
Where $w$ and $h$ denote the height and width of the images, respectively. 

\textbf{(2) F-Measure ($F_m$)} calculates a weighted harmonic mean of precision and recall. An adaptive thresholding approach is employed to establish the comparison of the binary saliency map with ground truth, wherein the threshold is set dynamically based on the saliency map. $F_{\beta}=(1+\beta^2)\cdot precision\cdot recall/(\beta^2\cdot precision+recall)$.

\textbf{(3) Weighted F-Measure ($W_F$)} is the F-measure with weighted precision (a measure of exactness) and weighted recall (a measure of completeness): $F_{\beta}^w=(1+\beta^2)\cdot precision^w\cdot recall^w/(\beta^2\cdot precision^w+ recall^w)$

\textbf{(4) Structure Measure ($S_m$}) assesses the structural integrity of the detected salient regions by incorporating two distinct components, region-aware ($S_r$) and object-aware ($S_o$). Unlike the MAE and F-measure, which compare two images on a pixel-by-pixel basis, the S-measure emphasises the structural similarity between the saliency map and ground truth: $S_{\alpha}=\alpha \cdot S_o+(1-\alpha)\cdot S_r$

\textbf{(5) E-Measure ($E_m$)} is the enhanced-alignment measure that combines local pixel values with the image-level mean value, jointly capturing image-level statistics and local pixel-matching information: $E_{\xi}=\sum_{i=1}^w\sum_{j=1}^h \varphi(i,j)/(w\times h)$. 
$\varphi$ represents the enhanced alignment matrix, which denotes the correlational relationship between the predicted saliency maps and the corresponding ground truth.

\subsection{Implementation Details}

The proposed model CSDNet is built and trained using the Pytorch framework. Optimisation during training is achieved through the application of the Adam optimisation algorithm. The MobileNet-V2 architecture, serving as the backbone of the network, is initialised with pre-trained weights obtained from the ImageNet dataset during the SAMAEP phase, whereas initialisation for the remaining network components is conducted randomly. All experimental training and testing procedures are performed on a machine equipped with an Intel 8C16T Core i7-11700KF at 3.6 GHz $\times$ 16 and an NVIDIA GeForce RTX 3060 graphics card.

\subsection{Experimental Results}

\begin{figure}[!ht]
    \centering
    \includegraphics[width=12cm]{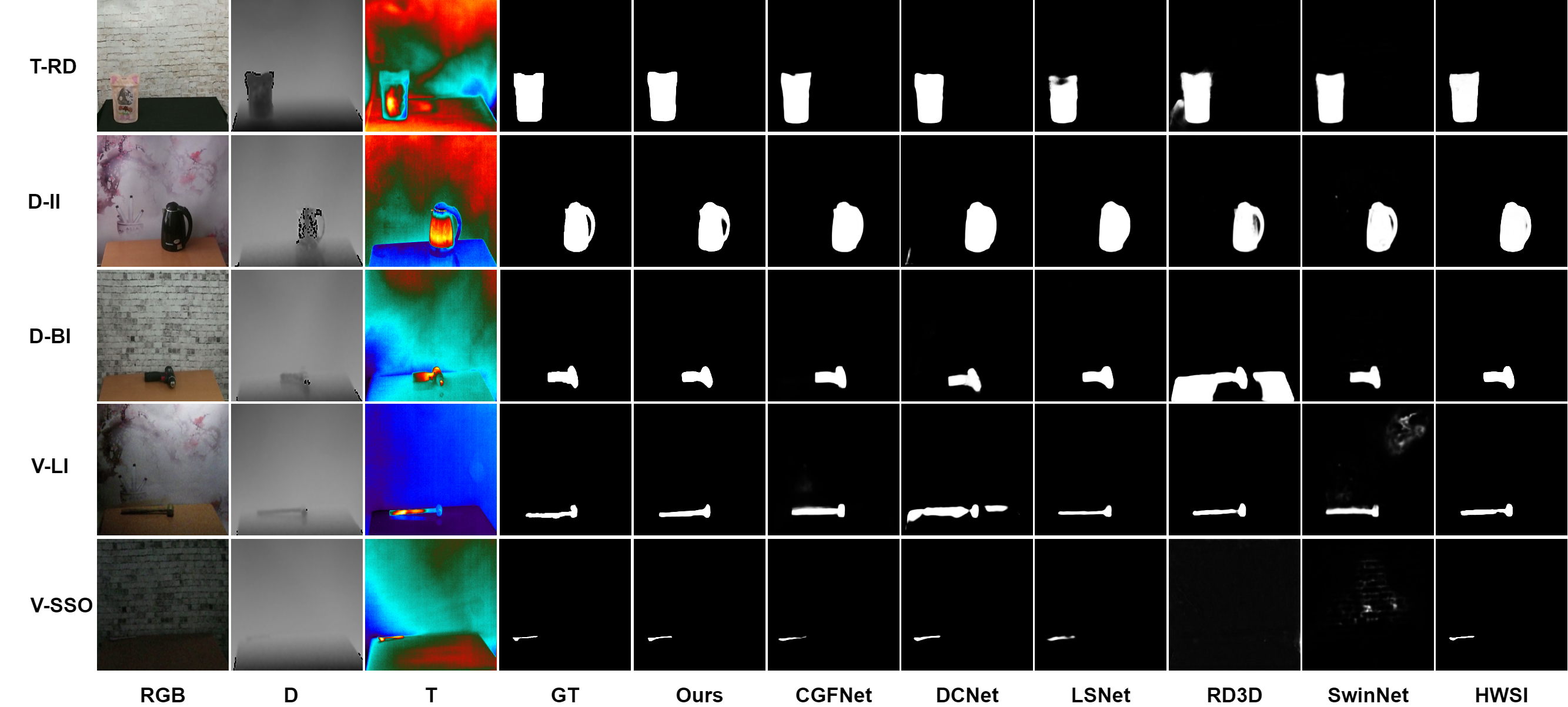}
    \caption{Visual Comparison on VDT-2048 dataset}
    \label{fig:vis-comp}
\end{figure}

We conduct the quantitative comparison of our proposed method with eight SOTA methods, including four RGB-T methods CGFNet \cite{wang2021cgfnet}, CSRNet \cite{huo2021efficient}, DCNet \cite{tu2022weakly}, LSNet \cite{zhou2023lsnet}, and four RGB-D methods MoADNet \cite{jin2022moadnet}, RD3D \cite{chen202233d}, SwinNet \cite{liu2021swinnet} and RFNet \cite{wu2022robust}. In addition, we include the RGB-D-T benchmark method HWSI \cite{song2022novel} for a comprehensive evaluation. To ensure a fair comparison, we use the official code released by the authors. 
The quantitative results are presented in Table \ref{tab:quan_vdt2048}, where the red colour highlights the best results among all double-modality methods, and bold text highlights the top-performing results across all types of methods. The proposed method demonstrates significant advantages over all other dual-modality approaches, regardless of whether they are RGB-D or RGB-T, and it achieves comparable results to the triple-modality benchmark method HWSI on the VDT-2048 dataset. Specifically, our method outperforms the other dual-modality approaches up to 0.97\%, 31.51\%, 30.37\%, 10.97\% and 14.61\% in terms of MAE, $F_m$, $W_F$, $S_m$ and $E_m$, respectively. Compared to the triple-modality method, the proposed method exhibits a mere 0.02\% disparity in MAE and gains 3.23\%, 0.41\% advantages in $F_m$ and $E_m$. For the remaining two metrics, $W_F$ and $S_m$, our method also achieves comparable results. The visual comparison is shown in Figure \ref{fig:vis-comp}. Furthermore, we analyse the triple-modality benchmark method HWSI using different bimodal inputs, demonstrating the comprehensive advantages of our method over the HSWI across various dual-modality combinations. The proposed method exhibits advantages up to 0.21\%, 5.61\%, 7.45\%, 3.95\% and 2.58\% in terms of the five SOD metrics, and the numerical comparison is detailed in Table \ref{tab:hswi-comp}. 

\begin{figure}[!ht]
    \centering
    \includegraphics[width=12cm]{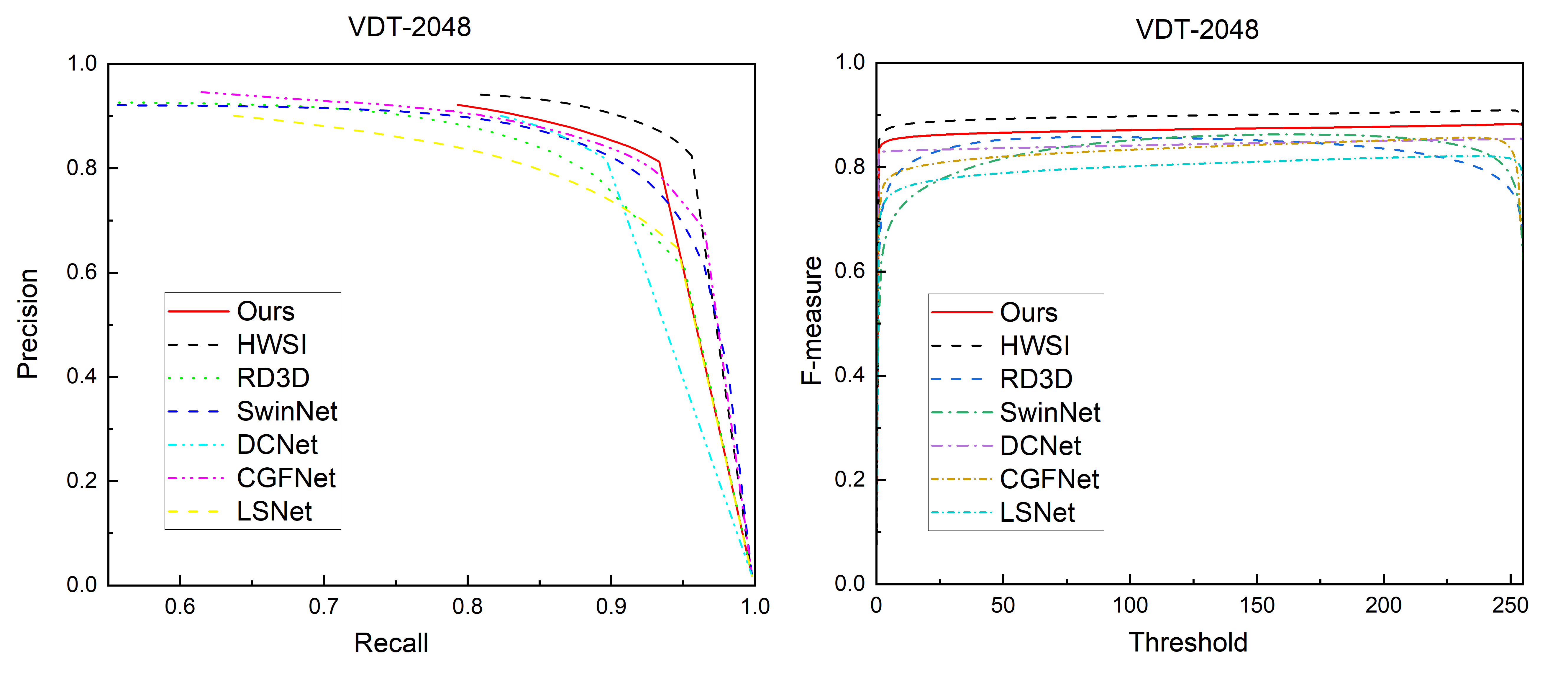}
    \caption{Precision-Recall Curve and $F_m$-Threshold Curve Comparison with Different Methods}
    \label{fig:pr-fm}
\end{figure}

\begin{table}[!ht]
\renewcommand{\arraystretch}{1.4}
\setlength{\tabcolsep}{3pt}
\caption{Quantitative Comparison Results of Different Methods on VDT-2048 Dataset. $\uparrow$/$\downarrow$ indicates that a larger/smaller value is better. The red text highlights the top result in dual-modality approaches, and the bold text highlights the top among all types of methods.}
\centering
\scalebox{0.9}{
\begin{tabular}{c|c|ccccc}
\hline
\hline
Model & Type & MAE$\downarrow$ & $F_m\uparrow$ & $W_F\uparrow$ & $S_m\uparrow$ & $E_m\uparrow$ \\
\hline
CGFNet & RGB-T & 0.0034 & 0.7777 & 0.8468 & 0.9166 & 0.9299      \\
CSRNet & RGB-T & 0.0050 & 0.7828 & 0.8159 & 0.8827 & 0.9460      \\
DCNet & RGB-T & 0.0038 & 0.8457 & 0.8284 & 0.8803 & 0.9699      \\
LSNet & RGB-T & 0.0045 & 0.7434 & 0.8046 & 0.8878 & 0.9201      \\
\hline
MoADNet & RGB-D & 0.0126 & 0.5753 & 0.5796 & 0.7697 & 0.8376      \\
RD3D & RGB-D & 0.0047 & 0.6444 & 0.7948 & 0.9090 & 0.8345      \\
SwinNet & RGB-D & 0.0038 & 0.7287 & 0.8385 & 0.9194 & 0.8962      \\
RFNet & RGB-D & 0.0031 & 0.8252 & 0.8680 & 0.9175 & 0.9635      \\
\hline
Ours & D-T & \textcolor{red}{0.0029} & \textbf{\textcolor{red}{0.8904}} & \textcolor{red}{0.8833} & 0.8794 & \textbf{\textcolor{red}{0.9806}}    \\
\hline
HWSI & RGB-D-T & \textbf{0.0027} & 0.8581 & \textbf{0.8983} & \textbf{\textcolor{red}{0.9324}} & 0.9765      \\
\hline
\hline
\end{tabular}
}
\label{tab:quan_vdt2048}
\end{table}
\begin{table}[!ht]
\renewcommand{\arraystretch}{1.4}
\setlength{\tabcolsep}{3pt}
\caption{Quantitative Comparison with HSWI on Different Modality Combinations}
    \centering
    \scalebox{0.9}{
    \begin{tabular}{c|c|ccccc}
    \hline
    \hline
     Model  &  Modality Settings & MAE$\downarrow$ & $F_m\uparrow$ & $W_F\uparrow$ & $S_m\uparrow$ & $E_m\uparrow$  \\
     \hline
      \multirow{3}{*}{HWSI}  &  RGB-D &  0.0048  &  0.8454  &  0.8179  &  0.8398  &  0.9548   \\
        &  RGB-T &  0.0034  &  0.8888  &  0.8708  &  0.8709  &  0.9653   \\
        &  D-T &  0.0050  &  0.8343  &  0.8088  &  0.8516  &  0.9580   \\
      \hline  
      Ours  &  D-T &  \textcolor{red}{0.0029}  &  \textcolor{red}{0.8904}  &  \textcolor{red}{0.8833}  &  \textcolor{red}{0.8793}  &  \textcolor{red}{0.9806}   \\
      \hline
      \hline
    \end{tabular}
    }
    \label{tab:hswi-comp}
\end{table}
In Figure \ref{fig:pr-fm}, our method is positioned in the upper-right corner among the PR curves and towards the top of the F-measure against the threshold diagram, demonstrating its superior performance. 
Moreover, to demonstrate the effectiveness and robustness of the proposed method, we also present the numerical results comparison with other methods on the challenges proposed in the VDT-2048 dataset. We assess the V-challenges encompassing low illumination (LI), no illumination (NI), side illumination (SI), small salient objects (SSO) in Table \ref{tab:v-challenge}, D-challenges including background interference (BI), background messy (BM), and information incomplete (II) in Table \ref{tab:d-challenge}, and T-challenges including thermal crossover (Cr), heat reflection (HR), and radiation dispersion (RD) in Table \ref{tab:t-challenge}. Since the proposed method is designed for indoor privacy-preserving applications and mobile platforms in search and rescue operations, it does not rely on the RGB visible light data. Thus, our approach exhibits significant advantages in all challenging illumination scenarios. However, in the SSO challenge, our method trails the HWSI by 0.04\%, 7.53\%, 1.2\% in terms of MAE, $W_F$, $E_m$ respectively, since the triple-modality does have a much richer texture that can benefit the segmentation in small targets. Despite this, our method, which emphasises the synergy between depth and thermal data by introducing CFARSP and ICAN, outperforms most dual-modality methods that primarily use visible light data and achieve comparable results to the trimodal HWSI. 
\begin{table}[ht]
\renewcommand{\arraystretch}{1.3}
\caption{Quantitative Results in V Challenges. LI, NI, SI, and SSO denote Low Illumination, No Illumination, Side Illumination and Small Salient Object, respectively}
    \centering
    \scalebox{0.8}{
    \begin{tabular}{c|ccc|ccc|ccc|ccc}
    \hline
    \hline
       \multirow{2}{*}{Model}  &   \multicolumn{3}{|c|}{V-LI}  &  \multicolumn{3}{|c|}{V-NI}  &  \multicolumn{3}{|c|}{V-SI}  &  \multicolumn{3}{|c}{V-SSO}  \\
         &     MAE$\downarrow$ & $W_F\uparrow$ & $E_m\uparrow$ &  MAE$\downarrow$ & $W_F\uparrow$ & $E_m\uparrow$ &  MAE$\downarrow$ & $W_F\uparrow$ & $E_m\uparrow$ &  MAE$\downarrow$ & $W_F\uparrow$ & $E_m\uparrow$   \\
    \hline
    CGFNet & .0046 & .8287 & .9334 & .0035 & .7614 & .8670 & .0043 & .8404 & .9324 & .0012 & .7196 & .7884 \\
    CSRNet & .0058 & .8117 & .9533 & .0043 & .7647 & .9185 & .0078 & .7905 & .9380 & .0014 & .7067 & .8188 \\
    DCNet  & .0048 & .8198 & .9474 & .0038 & .7291 & .9339 & .0047 & .8411 & \textbf{\textcolor{red}{.9833}} & .0012 & .6769 & \textbf{\textcolor{red}{.9324}} \\
    MoADNet & .0153 & .5543 & .8555 & .0143 & .3209 & .7917 & .0142 & .5776 & .8580 & .0049 & .4077 & .6391 \\
    PANet  & .1000 & .1360 & .7603 & .1207 & .0927 & .6804 & .1001 & .1271 & .7526 & .1321 & .0152 & .3773 \\
    LSNet  & .0057 & .7895 & .9297 & .0053 & .6734 & .8634 & .0064 & .7721 & .9301 & .0018 & .6491 & .7245 \\
    RD3D   & .0062 & .7583 & .8258 & .0066 & .6110 & .6964 & .0070 & .7521 & .8224 & .0025 & .6646 & .5698 \\
    SwinNet & .0050 & .8059 & .8864 & .0055 & .6718 & .7717 & .0055 & .8075 & .8862 & .0016 & .7039 & .6073 \\
    RFNet  & .0043 & .8450 & .9633 & .0043 & .7292 & .9161 & .0053 & .8179 & .9513 & \textcolor{red}{.0011} & \textcolor{red}{.7858} & .8816 \\
    \hline
    Ours   & \textbf{\textcolor{red}{.0031}} & \textbf{\textcolor{red}{.8927}} & \textcolor{red}{.9674} & \textbf{\textcolor{red}{.0024}} & \textbf{\textcolor{red}{.8610}} & \textbf{\textcolor{red}{.9541}} & \textbf{\textcolor{red}{.0033}} & \textbf{\textcolor{red}{.8934}} & .9740 & .0012 & .7649 & .9254 \\
    \hline
    HWSI   & .0038 & .8683 & \textbf{.9695} & .0028 & .8453 & .9522 & .0038 & .8757 & .9734 & \textbf{.0008} & \textbf{.8402} & .9134 \\ 
    \hline
    \hline
    \end{tabular}
    }
    \label{tab:v-challenge}
\end{table}
\begin{table}[!ht]
\renewcommand{\arraystretch}{1.3}
\setlength{\tabcolsep}{3pt}
\caption{Quantitative Results in D-Challenges. BI, BM and II Denote Background Interference, Background Messy and Information Incomplete Respectively}
    \centering
    \scalebox{0.8}{
    \begin{tabular}{c|ccc|ccc|ccc}
    \hline
    \hline
    \multirow{2}{*}{Model}  &   \multicolumn{3}{|c|}{D-BI}  &  \multicolumn{3}{|c|}{D-BM}  &  \multicolumn{3}{|c}{D-II}  \\
         &     MAE$\downarrow$ & $F_m\uparrow$ & $E_m\uparrow$ &  MAE$\downarrow$ & $F_m\uparrow$ & $E_m\uparrow$ &  MAE$\downarrow$ & $F_m\uparrow$ & $E_m\uparrow$    \\
    \hline
    MoADNet & .0107 & .5750 & .8353 & .0147 & .4674 & .8017 & .0183 & .5788 & .8461 \\
    PANet   & .1084 & .4528 & .6916 & .1062 & .5024 & .7368 & .1013 & .6426 & .8271 \\
    RD3D    & .0046 & .6161 & .8167 & .0044 & .6383 & .8289 & .0052 & .7316 & .8901 \\
    SwinNet & .0036 & .6858 & .8650 & .0044 & .6958 & .8717 & .0047 & .7845 & .9205 \\
    RFNet   & .0029 & .8124 & .9596 & .0032 & .8130 & .9614 & .0037 & .8652 & .9752 \\
    \hline
    Ours    & \textcolor{red}{.0027} & \textbf{\textcolor{red}{.8808}} & \textbf{\textcolor{red}{.9789}} & \textcolor{red}{.0031} & \textbf{\textcolor{red}{.8909}} & \textbf{\textcolor{red}{.9807}} & \textbf{\textcolor{red}{.0035}} & \textbf{\textcolor{red}{.9200}} & \textbf{\textcolor{red}{.9859}} \\
    \hline
    HWSI    & \textbf{.0024} & .8501 & .9748 & \textbf{.0029} & .8460 & .9727 & \textbf{.0035} & .8829 & .9813 \\
    \hline
    \hline
    \end{tabular}
    }
    \label{tab:d-challenge}
\end{table}
\begin{table}[!ht]
\renewcommand{\arraystretch}{1.3}
\setlength{\tabcolsep}{3pt}
\caption{Quantitative Results in T-Challenges. Cr, HR, RD Represent Crossover, Heat Reflection and Radiation Dispersion Respectively}
    \centering
    \scalebox{0.8}{
    \begin{tabular}{c|ccc|ccc|ccc}
    \hline
    \hline
    \multirow{2}{*}{Model}  &   \multicolumn{3}{|c|}{T-Cr}  &  \multicolumn{3}{|c|}{T-HR}  &  \multicolumn{3}{|c}{T-RD}  \\
         &     MAE$\downarrow$ & $F_m\uparrow$ & $E_m\uparrow$ &  MAE$\downarrow$ & $F_m\uparrow$ & $E_m\uparrow$ &  MAE$\downarrow$ & $F_m\uparrow$ & $E_m\uparrow$    \\
    \hline
    CCGFNet & .0034 & .7350 & .9075 & .0029 & .8329 & .9668 & .0046 & .8522 & .9750 \\
    CSRNet  & .0050 & .7304 & .9234 & .0034 & .8495 & .9792 & .0061 & .8453 & .9769 \\
    LSNet   & .0042 & .7119 & .8986 & .0043 & .7826 & .9489 & .0065 & .8083 & .9647 \\
    DCNet   & .0039 & .8116 & .9548 & .0032 & .8928 & \textbf{\textcolor{red}{.9893}} & .0051 & .8845 & \textcolor{red}{.9845} \\
    \hline
    Ours    & \textcolor{red}{.0032} & \textbf{\textcolor{red}{.8622}} & \textcolor{red}{.9558} & \textbf{\textcolor{red}{.0022}} & \textbf{\textcolor{red}{.9459}} & .9780 & \textbf{\textcolor{red}{.0037}} & \textbf{\textcolor{red}{.9202}} & .9732 \\
    \hline
    HWSI    & \textbf{.0025} & .8323 & \textbf{.9670} & .0027 & .8818 & .9878 & .0041 & .8854 & \textbf{.9888} \\
    \hline
    \hline
    \end{tabular}
    }
    \label{tab:t-challenge}
\end{table}
Table \ref{tab:flops-fps-para} list the quantitative comparison of the proposed method against the other SOTA methods in terms of running time, number of parameters and FLOPs. The proposed model achieves comparable performance with HWSI while utilising only 0.06 times the number of parameters and requiring 0.0036 times fewer FLOPs. In terms of processing speed, our model is approximately 5.97 times faster than HWSI. These results demonstrate that our model is more suited for deployment on edge devices or mobile platforms. 
\begin{table}[!ht]
\renewcommand{\arraystretch}{1.3}
\setlength{\tabcolsep}{3pt}
\caption{Comparison in terms of running time, model parameters and FLOPs}
    \centering
    \scalebox{0.9}{
    \begin{tabular}{c|ccccccc}
    \hline
    \hline
    Model  & CGFNet & DCNet & LSNet & RD3D & SwinNet & HWSI & Ours \\
    \hline
    Runtime (FPS) & 3.59 & 6.15 & 12.83 & 9.03 & 6.10 & 1.72 & 10.27  \\
    Params (M) & 66.38 & 24.06 & 4.56 & 46.90 & 198.78 & 100.77 & 5.96   \\
    FLOPs (G) & 345.54 & 207.21 & 1.23 & 50.86 & 124.72 & 357.93 & 1.30   \\
    \hline
    \hline
    \end{tabular}}
    \label{tab:flops-fps-para}
\end{table}
\subsection{Ablation Analysis}
To demonstrate the effectiveness of each module in our CSDNet (\ie, CFARSP, ICAN and SAMAEP), we conduct the ablation experiments and all the numerical results are presented in Table \ref{tab:ablation}. The results demonstrate that each module contributes to improving the model performance. It also indicates that the proposed method achieves the effective integration of depth and thermal modalities based on the intrinsic characteristics of the modality. 
\begin{table}[!ht]
\renewcommand{\arraystretch}{1.3}
\setlength{\tabcolsep}{3pt}
\caption{Ablation Study on Different Modules}
    \centering
    \scalebox{0.9}{
    \begin{tabular}{c|ccccc}
    \hline
    \hline
    Settings & MAE$\downarrow$ & $F_m\uparrow$ & $W_F\uparrow$ & $S_m\uparrow$ & $E_m\uparrow$  \\
    \hline
     w/o CFARSP  & 0.0033 & 0.8815 & 0.8727 & 0.8758 & 0.9745  \\
     w/o ICAN    & 0.0033 & 0.8810 & 0.8664 & 0.8748 & 0.9758  \\
     w/o SAMAEP  & 0.0031 & 0.8892 & 0.8813 & 0.8793 & 0.9783  \\
     Ours        & 0.0029 & 0.8904 & 0.8833 & 0.8793 & 0.9806  \\
     \hline
     \hline
    \end{tabular}}
    \label{tab:ablation}
\end{table}
\section{Conclusion}

In this study, we present a novel lightweight cross shallow and deep perception network, CSDNet, to effectively integrate different modalities with low coherence. The proposed method is implemented and assessed on salient object detection task with depth and thermal imagery. SAMAEP employs SAM to guide the encoder in learning the way to map features to a more generalised feature space. Our method, which leverages the depth-thermal modality, outperforms the current SOTA RGB-D and RGB-T approaches and achieves comparable performance with RGB-D-T method on VDT-2048 dataset. It runs 5.96 times faster and requires 0.0036 times fewer FLOPs than the triple-modality method, making it suitable for deployment in edge device applications with privacy concerns (\eg home care) or mobile platforms operating under challenging lighting conditions (\eg search and rescue robots). Extensive experiments are carried out under different challenging conditions, including difficult illuminations, small objects, background interference and various thermal interferences, to demonstrate the robustness of our method. CSDNet demonstrates effective integration of low-coherence depth-thermal modality and shows great potential to be generalised to the other modalities with low-coherence.

%
%

\end{document}